\documentclass[10pt,journal,compsoc]{IEEEtran}
\ifCLASSOPTIONcompsoc
  \usepackage[nocompress]{cite}
\else
  \usepackage{cite}
\fi

\usepackage{subfig}
\ifCLASSINFOpdf
\else
\fi

\usepackage{graphicx,multirow,float,epstopdf,amssymb,amsmath,xcolor,caption}
\usepackage{array}

\begin{document}
%
% paper title
% Titles are generally capitalized except for words such as a, an, and, as,
% at, but, by, for, in, nor, of, on, or, the, to and up, which are usually
% not capitalized unless they are the first or last word of the title.
% Linebreaks \\ can be used within to get better formatting as desired.
% Do not put math or special symbols in the title.
\title{Joint Face Completion and Super-resolution using Multi-scale Feature Relation Learning}
%
%
% author names and IEEE memberships
% note positions of commas and nonbreaking spaces ( ~ ) LaTeX will not break
% a structure at a ~ so this keeps an author's name from being broken across
% two lines.
% use \thanks{} to gain access to the first footnote area
% a separate \thanks must be used for each paragraph as LaTeX2e's \thanks
% was not built to handle multiple paragraphs
%
%
%\IEEEcompsocitemizethanks is a special \thanks that produces the bulleted
% lists the Computer Society journals use for "first footnote" author
% affiliations. Use \IEEEcompsocthanksitem which works much like \item
% for each affiliation group. When not in compsoc mode,
% \IEEEcompsocitemizethanks becomes like \thanks and
% \IEEEcompsocthanksitem becomes a line break with idention. This
% facilitates dual compilation, although admittedly the differences in the
% desired content of \author between the different types of papers makes a
% one-size-fits-all approach a daunting prospect. For instance, compsoc
% journal papers have the author affiliations above the "Manuscript
% received ..."  text while in non-compsoc journals this is reversed. Sigh.

\author{Zhilei~Liu,
        Yunpeng~Wu,
        Le~Li,
        Cuicui~Zhang,
        and~Baoyuan~Wu
\IEEEcompsocitemizethanks{
\IEEEcompsocthanksitem Corresponding author: Cuicui~Zhang.
\IEEEcompsocthanksitem Z. Liu, Y. Wu, and L. Li are with the College of Intelligence and Computing, Tianjin University, Tianjin 300072, China. E-mail: {Zhileiliu, wuyunpeng, le\_li}@tju.edu.cn.
\IEEEcompsocthanksitem C. Zhang is with the School of Marine Science and Technology, Tianjin University, Tianjin, 300072, China. E-mail: cuicui.zhang@tju.edu.cn.
\IEEEcompsocthanksitem B. Wu is with the Tencent AI Lab, Shenzhen, China, 518057. E-mail: wubaoyuan1987@gmail.com.
}% <-this % stops an unwanted space
\thanks{Manuscript received February, 2020.}}

% note the % following the last \IEEEmembership and also \thanks -
% these prevent an unwanted space from occurring between the last author name
% and the end of the author line. i.e., if you had this:
%
% \author{....lastname \thanks{...} \thanks{...} }
%                     ^------------^------------^----Do not want these spaces!
%
% a space would be appended to the last name and could cause every name on that
% line to be shifted left slightly. This is one of those "LaTeX things". For
% instance, "\textbf{A} \textbf{B}" will typeset as "A B" not "AB". To get
% "AB" then you have to do: "\textbf{A}\textbf{B}"
% \thanks is no different in this regard, so shield the last } of each \thanks
% that ends a line with a % and do not let a space in before the next \thanks.
% Spaces after \IEEEmembership other than the last one are OK (and needed) as
% you are supposed to have spaces between the names. For what it is worth,
% this is a minor point as most people would not even notice if the said evil
% space somehow managed to creep in.

% The paper headers
%Journal of \LaTeX\ Class Files,~Vol.~14, No.~8, August~2015
\markboth{IEEE Transactions on XXX,~Vol.~X,~NO.~X,~X}%
{Shell \MakeLowercase{\textit{et al.}}: Bare Demo of IEEEtran.cls for Computer Society Journals}
% The only time the second header will appear is for the odd numbered pages
% after the title page when using the twoside option.
%
% *** Note that you probably will NOT want to include the author's ***
% *** name in the headers of peer review papers.                   ***
% You can use \ifCLASSOPTIONpeerreview for conditional compilation here if
% you desire.

% The publisher's ID mark at the bottom of the page is less important with
% Computer Society journal papers as those publications place the marks
% outside of the main text columns and, therefore, unlike regular IEEE
% journals, the available text space is not reduced by their presence.
% If you want to put a publisher's ID mark on the page you can do it like
% this:
%\IEEEpubid{0000--0000/00\$00.00~\copyright~2015 IEEE}
% or like this to get the Computer Society new two part style.
%\IEEEpubid{\makebox[\columnwidth]{\hfill 0000--0000/00/\$00.00~\copyright~2015 IEEE}%
%\hspace{\columnsep}\makebox[\columnwidth]{Published by the IEEE Computer Society\hfill}}
% Remember, if you use this you must call \IEEEpubidadjcol in the second
% column for its text to clear the IEEEpubid mark (Computer Society jorunal
% papers don't need this extra clearance.)

% use for special paper notices
%\IEEEspecialpapernotice{(Invited Paper)}

% for Computer Society papers, we must declare the abstract and index terms
% PRIOR to the title within the \IEEEtitleabstractindextext IEEEtran
% command as these need to go into the title area created by \maketitle.
% As a general rule, do not put math, special symbols or citations
% in the abstract or keywords.
\IEEEtitleabstractindextext{%
\begin{abstract}
Previous research on face restoration often focused on repairing a specific type of low-quality facial images such as low-resolution (LR) or occluded facial images. However, in the real world, both the above-mentioned forms of image degradation often coexist. Therefore, it is important to design a model that can repair LR occluded images simultaneously. This paper proposes a multi-scale feature graph generative adversarial network (MFG-GAN) to implement the face restoration of images in which both degradation modes coexist, and also to repair images with a single type of degradation. Based on the GAN, the MFG-GAN integrates the graph convolution and feature pyramid network to restore occluded low-resolution face images to non-occluded high-resolution face images. The MFG-GAN uses a set of customized losses to ensure that high-quality images are generated. In addition, we designed the network in an end-to-end format. Experimental results on the public-domain CelebA and Helen databases show that the proposed approach outperforms state-of-the-art methods in performing face super-resolution (up to 4x or 8x) and face completion simultaneously. Cross-database testing also revealed that the proposed approach has good generalizability.
\end{abstract}

% Note that keywords are not normally used for peerreview papers.
}

% make the title area
\maketitle

% To allow for easy dual compilation without having to reenter the
% abstract/keywords data, the \IEEEtitleabstractindextext text will
% not be used in maketitle, but will appear (i.e., to be "transported")
% here as \IEEEdisplaynontitleabstractindextext when the compsoc
% or transmag modes are not selected <OR> if conference mode is selected
% - because all conference papers position the abstract like regular
% papers do.
\IEEEdisplaynontitleabstractindextext
% \IEEEdisplaynontitleabstractindextext has no effect when using
% compsoc or transmag under a non-conference mode.

% For peer review papers, you can put extra information on the cover
% page as needed:
% \ifCLASSOPTIONpeerreview
% \begin{center} \bfseries EDICS Category: 3-BBND \end{center}
% \fi
%
% For peerreview papers, this IEEEtran command inserts a page break and
% creates the second title. It will be ignored for other modes.
\IEEEpeerreviewmaketitle

\IEEEraisesectionheading{\section{Introduction}\label{sec:introduction}}

\IEEEPARstart{G}{enerative} face restoration aims to recover valuable missing information of face images caused by factors such as low resolution (LR), occlusion, and large pose. It is the subject of extensive research in the field of face recognition, especially with the emergence of convolution neural networks (CNNs) \cite{simonyan2014very,krizhevsky2012imagenet},  and generative adversarial networks (GANs) \cite{goodfellow2014generative}. Many face image restoration sub-tasks have yielded great breakthroughs, including face completion \cite{pathak2016context,yeh2017semantic}, face super-resolution\cite{yang2019deep} or hallucination  \cite{chen2018fsrnet,ledig2017photo}, and face frontal view synthesis \cite{huang2017beyond}. 

\begin{figure}[!htbp] 
\centering 
\includegraphics[scale=0.32]{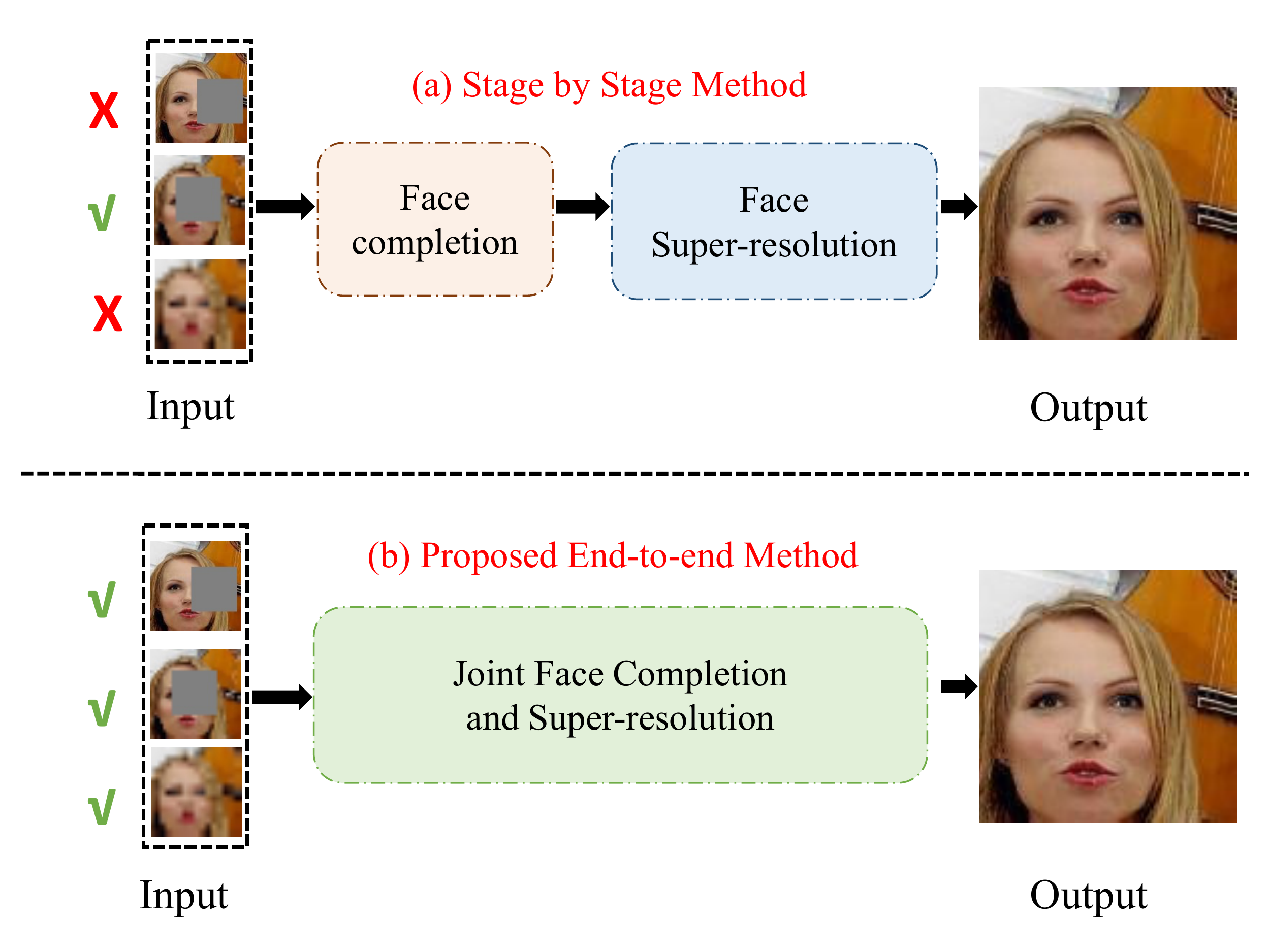} 
\caption{(a) Traditional two-stage multi-task face restoration model; (b) The proposed end-to-end face restoration model.} 
\label{fig:fig1}
\end{figure}

Although many methods have been proposed for image completion and image super-resolution reconstruction, most are designed for single tasks such as face completion \cite{li2017generative,song2018geometry} and face super-resolution \cite{cao2017attention,chen2018fsrnet,ledig2017photo,song2017learning,zhu2016deep}. Thus, these methods are more effective at single tasks; however, they underperform when applied to multiple tasks. Both occluded and LR pictures can be perceived as original pictures with degradation matrix added to them, which can be expressed as follows:
\begin{equation}
\label{eq:Joint}
\widehat{I}_{m\times n}=\begin{bmatrix}
\theta _{1,1} & ...  & \theta _{1,n}\\ 
... & \theta _{i,j}  & ... \\ 
\theta _{m,1} & ...  & \theta _{m,n} 
\end{bmatrix}_{N_{m*n}}\odot I_{m\times n}
\end{equation}
where $\hat I_{m*n}$ is a low-quality image, $I_{m*n}$ is an original image, and $N_{m*n}$ is a $m*n$ degeneration matrix. For occluded images, $\theta_{i,j}$ is 0 or 1; for LR images, $\theta_{i,j}$ is any real number between 0 and 1, and $\odot$ stands for element multiplication. Repairing both forms of degradation  requires eliminating the influence of the degradation matrix $N_{m*n}$ on the original image $I_{m*n}$. For both tasks, the most direct way to achieve this is to connect two pre-trained models in series. The output of the first model is inputted into the second model to repair the LR occluded pictures. However, this method tends to further increase the impact of noise in low-quality pictures, making the repair effect poor. Cai et al.~\cite{cai2019fcsr} proposed combining two pre-trained models and modifying the training strategy to achieve image completion and image super-resolution tasks simultaneously. Although this method has proven effective at processing multiple tasks, it cannot complete a single task because it relies on two pre-trained models. 

In this paper, we propose an end-to-end model that can not only achieve both image completion and image super-resolution simultaneously, but is also effective at a single task. The comparison of the two different models is shown in Fig.~\ref{fig:fig1}. Compared with the general model, our model can effect both multi-task repair and single-task repair. In addition, unlike existing face image restoration models, to learn the features of occluded face patches from the non-occluded patches, we propose an improved graph convolutional network (IGCN), whose structure is shown in Fig.~\ref{fig:fig3}. The IGCN can also improve the resolution of non-occluded face patches from other visible face patches by exploring the correlations of different facial components in different facial expressions. Then, using the proposed IGCN, a region relation modeling block (RRMB) is built for capturing facial features with different scales for face restoration, as shown in Fig.~\ref{fig:fig4}. Given finer facial division, the proposed framework can represent the relations among different face patches very accurately. Because of the very accurate adjacency matrix in the proposed IGCN, our model can effectively restore the feature map in deep networks using the features of the visible patches. Furthermore, to consider the relationship between different feature sizes in the face region, we incorporate the feature pyramid network (FPN) \cite{lin2017feature} into the model and use the pre-trained VGG-19 network as the feature extraction network in the FPN. Finally, we design a discriminator to enable the generator to generate realistic faces. The model includes three losses: (i) pixel loss for effective reconstruction of non-occluded high-resolution face images; (ii) adversarial loss for differentiating between real and generated face images; (iii) perceptual loss for improving the quality of the generated faces to some extent.

The main contributions of this work include the following: (i) An end-to-end model that can complete not only image completion and image super-resolution tasks but also single tasks; (ii) Unlike the existing model, based on the relationship between various regions of the image, we used the graph convolution method instead of the conventional convolution method. Apart from this, considering that different features have different sizes, we incorporated the FPN into the model; (iii) Compared with state-of-the-art face image completion and face image super-resolution methods, the results are promising. 

This work is an extension of our previous work on IGCN~\cite{liu2020facial}. The essential improvements over our previous work include the following: (i) Improvement on the model and incorporation of the FPN; (ii) Verification of the effectiveness of the model on single tasks; (iii) Extension of the datasets to the general face datasets.

\section{Related Work}

We will briefly describe the existing research on image restoration, as well as provide an introduction to FPN and graph convolutional networks (GCNs).

\subsection{Image Restoration}
\noindent
Image restoration consists of image super-resolution or hallucination \cite{chen2018fsrnet,ledig2017photo}, image completion \cite{li2017generative,yeh2017semantic}, face frontal view synthesis \cite{huang2017beyond}, image denoizing \cite{muhammad2018image}, image deraining \cite{wang2018rain}, image dehazing \cite{song2017single,yang2019single}, image deblurring \cite{li2018learning}, and shadow removal \cite{wang2018stacked}. In this study, we mainly addressed image completion and image super-resolution tasks.

In image completion, the entire area of the image is utilized to fill in the missing area. Early image completion methods typically utilized information from the surrounding pixels of the occluded area to recover the missing parts. Ballester et al. \cite{ballester2001filling} propose joint interpolation of the gray and gradient directions of the image to fill in the missing areas; however, this method is ineffective when the missing areas are large or have pixels with values varying significantly from those of the complete area. Hays J et al. \cite{hays2007scene} attempted to use the data-driven method to solve the shortcomings of the previous method. When no similar patch can be found in the visible parts of the image, this study suggests that material can be obtained from the large number of pictures on the Internet. To solve the problem of low overall efficiency, this article proposed the extraction of a complete block directly from other images to fill the hole. Efros et al. \cite{cai2019fcsr} propose a patch-based method to search for relevant patches from the non-occluded areas of the image, and use them to gradually fill in the missing areas from the outside to the inside. Although this method improves on the previous study, the area search process is time-consuming. To solve this problem, Barnes et al. \cite{barnes2009patchmatch} propose a fast patch search algorithm; however, their method cannot complete image completion in real time. With the development of deep learning, the application of deep models to face image restoration commenced. Liu et al. \cite{liu2018image} use CNNs to gradually recover lost pixels. Pathak D et al. \cite{pathak2016context} propose the use of CNNs for learning high-level features in images, following which these features are used to guide the process of generating the missing parts of images. Recently, image restoration has attracted significant attention, because the GAN \cite{goodfellow2014generative} can generate an image from a random vector, and use a discriminator to distinguish the real image from generated image \cite{chen2018fsrnet,song2018geometry}. The conditional GAN was proposed to limit the distribution of generated images \cite{mirza2014conditional}. 
In addition, some additional prior knowledge were also introduced into the repair process. Wei et al. \cite{xiong2019foreground} proposed to add the contour information of image as a constraint to the generation network to ensure the contour details of the generated image Many methods have been developed to tackle the task of image completion; however, they are generally not effective at multi-task processing, as has been demonstrated above. 

In image super-resolution, a set of LR images (or motion sequences) is used to generate a single high-resolution image. The application field of image super-resolution reconstruction is wide, and it has important application prospects in areas such as military, medicine, public safety, and computer vision. Before deep learning, learning-based super-resolution algorithms were a hot topic. Learning-based super-resolution algorithms use a large number of high-resolution (HR) image construction learning libraries to generate learning models. To repair LR images, the algorithms exploited prior knowledge obtained from the learning model to obtain the high-frequency details of the image, and achieve better image recovery effect. Jiang et al. \cite{jiang2016srlsp} regarded the face SR image as an image interpolation problem in a specific domain. An intensity missing interpolation method based on local prior smooth regression is proposed, referred to as SRLSP. The author assumed that facial image blocks at the same location share similar local structures and use smooth regression to understand the relationship between LR pixels and the missing HR pixels of a location block.

Deep learning was applied for the first time to image super-resolution by Dong et al. \cite{ren2016single}. They proposed a super-resolution CNN (SRCNN) method for image super-resolution. The main steps of the algorithm are as follows: 1) HR images were obtained from LR images. 2) images were obtained through neural network feature extraction, non-linear feature mapping, and reconstruction. Compared with the traditional method, this method greatly improved the reconstruction of HR images. Kim et al. \cite{kim2016deeply} propose a deeply-recursive convolutional network (DRCN) structure with a network that was deeper than that of the SRCNN. Although the CNN achieved some breakthroughs in the completion of HR images, it also had certain limitations. When the resolution of the picture is too low, the CNNs often cannot repair it effectively. Huang et al. \cite{huang2017wavelet} proposed a wavelet-based CNN method, Wavelet-SRNet, to process extremely LR pictures. The WT can describe the context and texture information of the image from different levels such that the repaired picture is closer to the real picture. Shi et al. \cite{shi2017structure} used a multi-task learning approach, first performed feature extraction on LR pictures and then applied a deconvolution module to interpolate LR feature maps in a content-adaptive manner. Then, the generated feature map was fed to the subnets of the two branches to retain as much high-frequency detail information as possible in the generated results. Other methods that repair extremely LR images to realize HR reconstruction by modifying the network structure have also been proposed. In 2017, Christian Ledig et al. propose the super-resolution GAN (SRGAN) \cite{ledig2017photo}. The authors pioneered the use of GAN to solve the problem of super-resolution. They mention that the mean square error is used as a loss function when the network is being trained; however, the recovered images usually lose high-frequency information, as a result of which the visual experience is reduced. The SRGAN used perceptual loss and adversarial loss to improve the realism of the recovered images. Although the recovered image has a lower peak signal-to-noise ratio (PSNR) value, it has realistic visual effects. Because of the significant effect of GAN in the generation field, more and more models upgraded on the basis of GAN are also widely used in the field of image super-resolution reconstruction. Guo et al.~\cite{guo2019auto} developed a novel GAN called Auto-Embedding Generative Adversarial Network (AEGAN), which simultaneously encodes the global structure features and captures the fine-grained details, learning from the idea of auto-encoder GAN. Similarly, the above methods focus on the response of single task.

\subsection{Feature Pyramid Networks}
\noindent
In 2017, Lin et al. \cite{lin2017feature} proposed the FPN, a multi-scale objection detection method. Most of the original object detection algorithms utilize only top-level features for prediction. Although the semantic information of the lower-level feature is relatively less, the target position is accurate. The high-level feature semantic information is rich; however, the target position is imprecise. In addition, although some algorithms use multi-scale feature fusion, they generally use the fused features for prediction. The uniqueness of the FPN is that the prediction is performed independently at different feature layers. The effectiveness of the FPN in the fields of object recognition and behavior recognition has popularized it greatly in the field of image restoration. In image super-resolution, Lai et al.~ \cite{lai2018fast} proposed the deep Laplacian pyramid super-resolution network for fast and accurate image super-resolution. This method greatly reduced the number of parameters, and saved running time. This method of introducing multi-layer features into image restoration also inspired people to introduce the FPN into image restoration. In image completion, Zeng et al.~\cite{zeng2019learning} proposed that the image be restored at the feature level and image level simultaneously. The restored low-level features enable the restoration of high-level features, and proved to be effective. In this study, to better retain the information of the face features, we also used the FPN as the feature extraction component of the network.

\subsection{Graph Convolutional Networks}
\noindent
Considering that there is a relationship between the domain to be repaired and the visible domain, we also applied the GCN \cite{zhou2018graph} to the overall experimental model. GCN-based methods can either be spectral-based or spatial-based. The spectral-based approach introduces filters from the perspective of graph signal processing to define graph convolution, where the graph convolution operation is interpreted as denoizing the graph signal. The spatial-based method represents the graph convolution as the aggregation of feature information from the neighborhood. When the algorithm of the graph convolution network is run at the node level, the graph pooling module and the graph convolution layer can be interleaved to coarsen the graph into an advanced substructure. Inspired by the first-order graph Laplacian methods, Kipf et al.~\cite{kipf2016semi} propose the GCN. A link matrix is defined according to the overall structure of the graph, and related nodes in the graph are connected through the defined link matrix to generate a new feature graph. Graph convolution takes into account the relationship between features, which also provides a theoretical basis for graph convolution rather than conventional convolution. Therefore, rather than utilize the linear transformation in the conventional GCN, we improved the conventional GCN by combining the tensor-inputs and the standard convolutional layer to retain the facial structure information.

\section{Proposed Method}
\noindent
In this section, the proposed method is described in details. First, the overall framework of our MFG-GAN, consisting of a generator and discriminator, is introduced in Section~\ref{sec:subsection3-1}. Then, in Section~\ref{sec:subsection3-2}, we introduce the three different losses used in our model.
\begin{figure*}[!htbp] 
\centering 
\includegraphics[width=\textwidth]{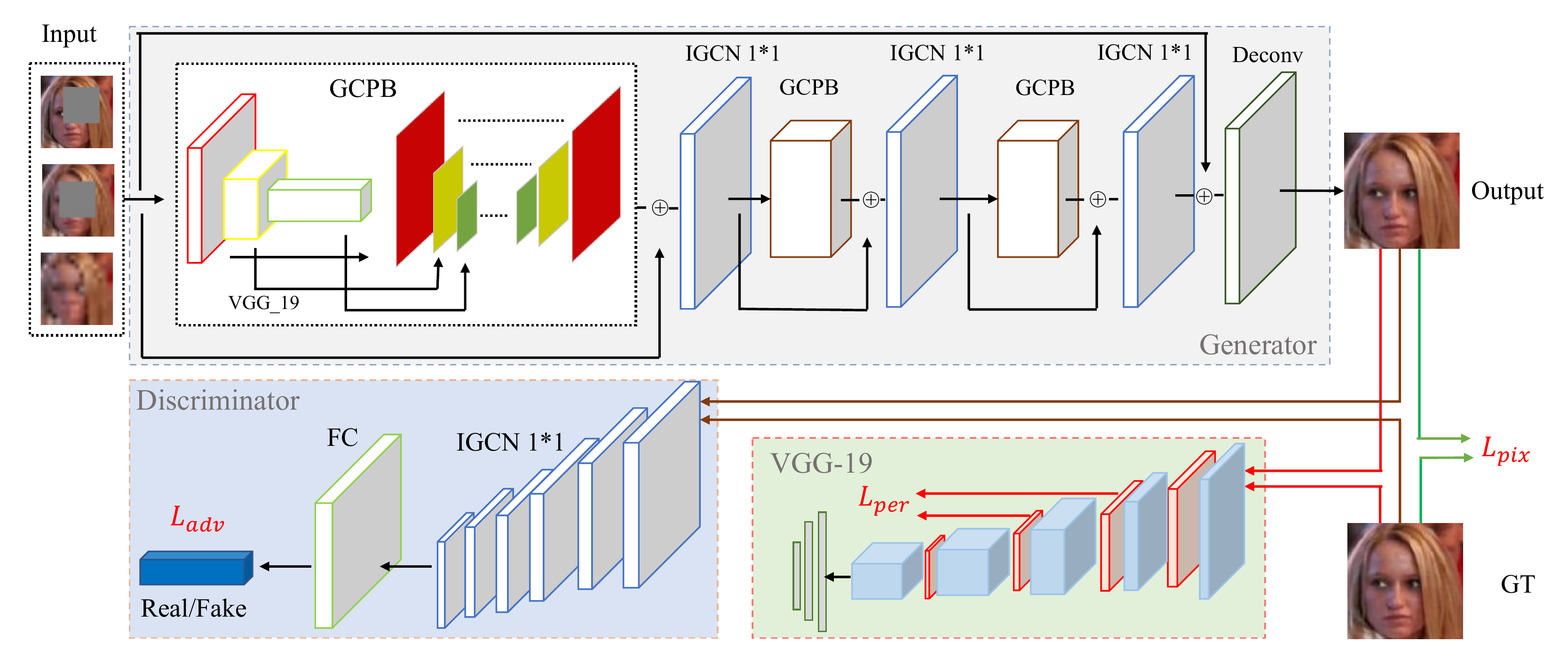} 
\caption{The overall framework of our proposed MFG-GAN model. The input is a low-quality image, which can be an occluded image, an LR image, or an LR occluded image. The generator includes three graph convolution pyramid blocks (GCPB), each of which is followed by an improved graph convolution layer; the last layer is the deconvolution layer; the discriminator is a six-layer improved graph convolution base layer and a fully connected layer that is used to distinguish the generated image from the real image. Besides, both perceptual loss from VGG-19 and pixel-wise loss are also considered.} 
\label{fig:fig2}
\end{figure*}

\subsection{Network Architecture}\label{sec:subsection3-1}
\begin{figure}
\centering
\includegraphics[scale=0.24]{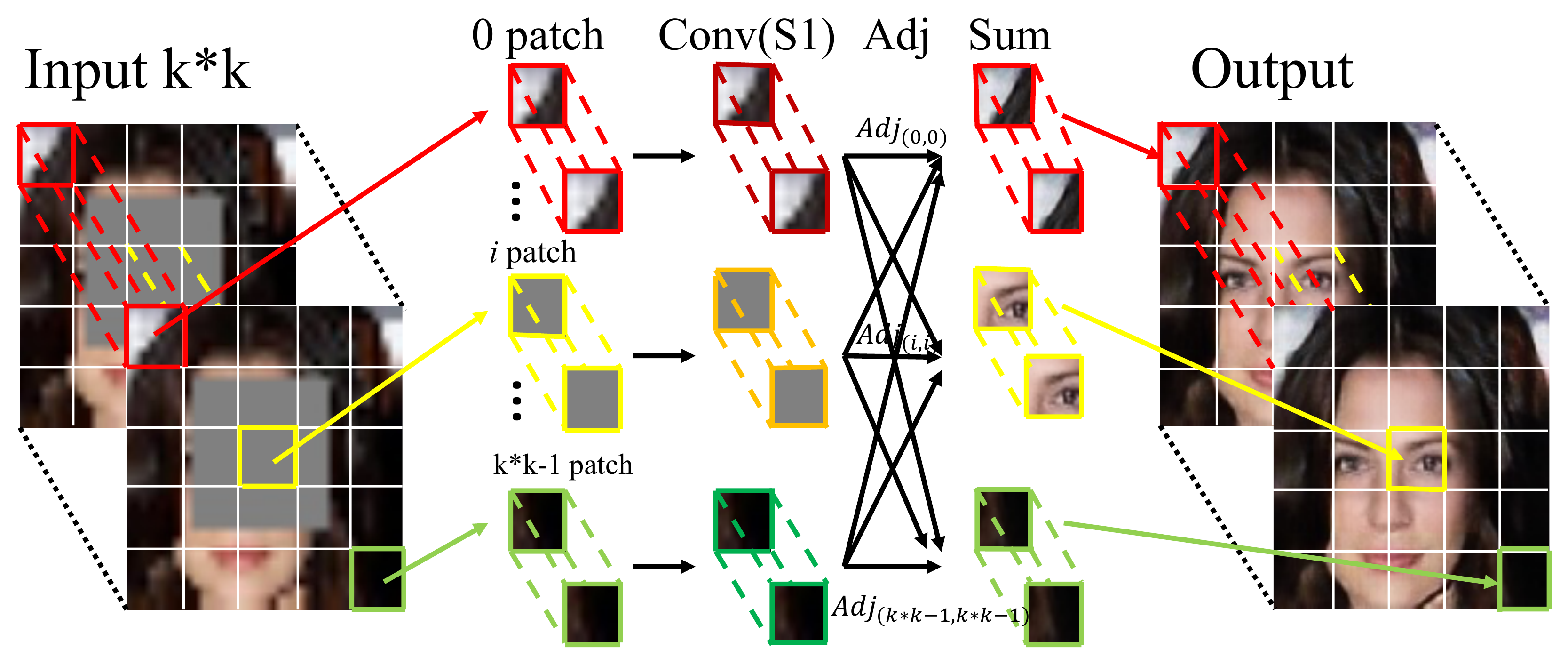}
\caption{Structure of the IGCN $k * k$ with stride 1. $Adj_{(i,j)}$ represents the link between $i$-th patch and $j$-th patch; $Adj$ represents adjacency matrix.}
\label{fig:fig3}
\end{figure}

\noindent
The structure of our proposed method is shown in Fig.~\ref{fig:fig2}; it consists of a generator to restore the entire face image and a discriminator to determine whether the generated face image is real or fake. To fully utilize the non-occluded face patches, we simultaneously addressed the face completion and super-resolution problems. For the face completion, we modeled the relations between the non-occluded face patches and occluded patches. For the face super-resolution, the correlation among different non-occluded face patches was modeled. The aim was to ensure the global harmony of the generated face images. In addition, we used graph convolution instead of general convolution. Unlike general graph convolution, we propose an improved GCN (IGCN). In the conventional graph convolutional network, the features of the nodes, which are referred to as non-Euclidean data, are vectors \cite{goodfellow2014generative}. Every patch of the face image is related to the other patches, and is Euclidean data. To directly utilize the face patches as nodes by modeling the relations among the different facial patches, we proposed an improved GCN, the entire structure of which is shown in Fig.~\ref{fig:fig3}. 

\begin{figure*}[!htp] 
    \centering
    \subfloat[Graph Convolution Pyramid Block (GCPB)]{%
        \includegraphics[width=0.5\textwidth]{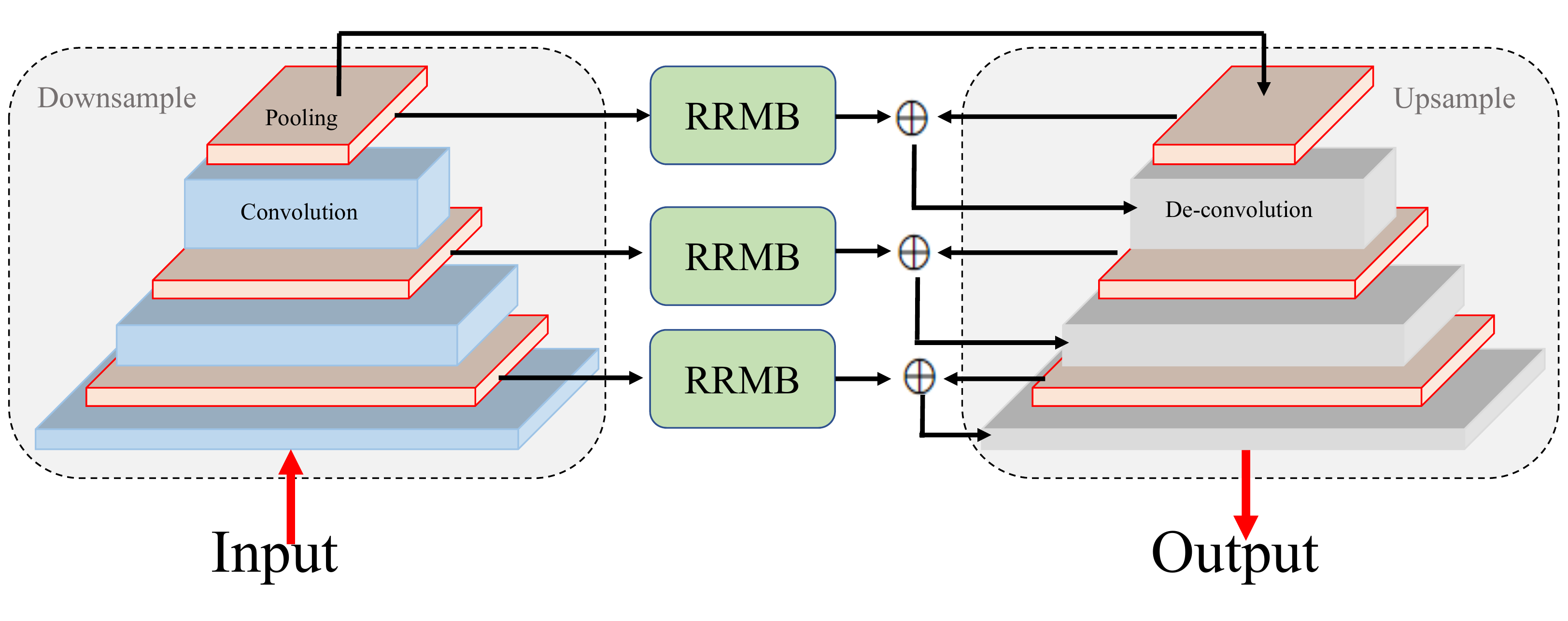}%
        \label{fig:fig4a}%
        }%
    \hfill%
    \subfloat[Region Relation Modeling Block (RRMB)]{%
        \includegraphics[width=0.45\textwidth]{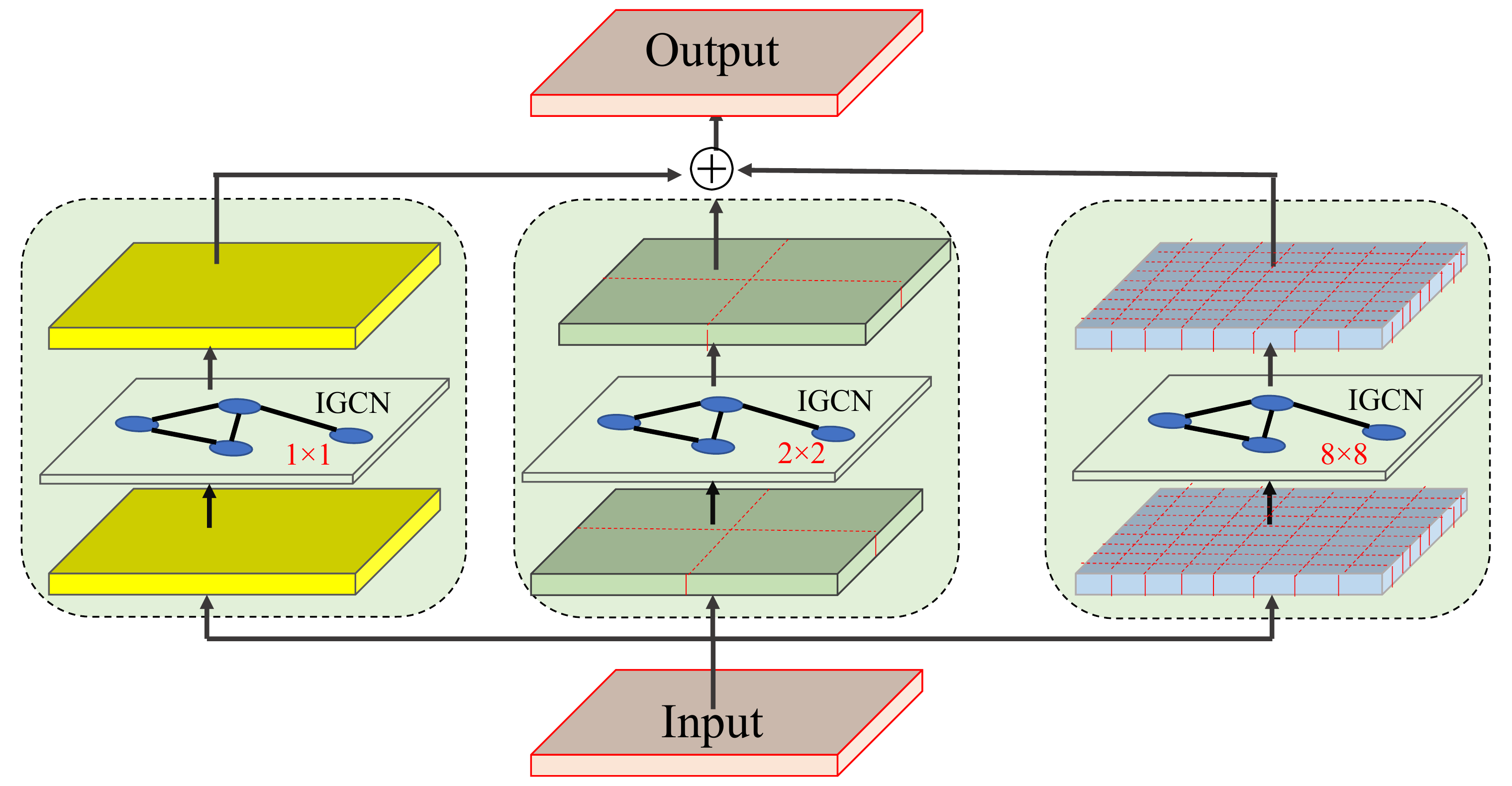}%
        \label{fig:fig4b}%
        }%
    \caption{Framework of GCPB and RRMB. (a) GCPB: Image is inputted into the VGG-19 network for feature extraction; the extracted features are inputted into the RRMB for graph convolution. Then, the convolved features are concatenated, and inputted into the corresponding deconvolutional layer. (b) RRMB consists of the IGCN $1 * 1$, $2 * 2$, $8 * 8$. $1 * 1$, is obtained by splitting only one patch for the inputs,  $2 * 2$ is obtained by splitting four patches for the inputs, and $8 * 8$ represents splitting 64 patches for the inputs.}
    \label{fig:fig4}
\end{figure*}

First, we split the feature, $F$, of the face image into $k * k$ face patches with specific order ID. We used the convolutional layer to capture the representation of each face patch. At this juncture, it should be noted that conventional GCNs deploy vectors as features, and use the linear transformation layer to capture representations. Unlike conventional GCNs, our proposed IGCN deployed the 4-D tensor of face patches as features, and used the convolution layer to capture representations. The weights of all the convolutional layers for each patch were distributed under one layer of the IGCN. According to the symmetrical adjacency matrix, we obtained every face patch feature after the summation operation. Finally, we converted the $k * k$ face patch features into a feature map according to the origin position ID. Note that the deconvolutional layer can also be used in the IGCN. The adjacency matrix was pre-defined using the facial structure. The IGCN can be defined as follows:
\begin{equation}
\label{eq_fupd}
F_{upd}=A:Relu(W\times F+b)
\end{equation}
where $F$ is the stacked patches features, $A$ is the normalized adjacency matrix, and b  represents the tensor product. The adjacency matrix is defined by the correlations among the various facial regions. Supposing that two patches are correlated, then, the link between the two patches ought to be 1, and the opposite, 0. The GCN makes it possible to use the non-occluded regions to complete the occluded regions via pre-defined relations; for instance, the non-occluded left eye can be used to restore the occluded right eyes. Likewise, the non-occluded region can be used to enhance the quality of other non-occluded regions. Besides, to reflect the relationship between different size features, we designed a GCPB to replace the general convolution layer; each GCPB is composed of a FPN and three RRMBs, as shown in Fig.~\ref{fig:fig4a}.

We used the pre-trained VGG-19 network as the feature extraction component of the FPN, sent the extracted features of the first three blocks to the RRMB for convolution, following which we concatenated the obtained features as a new feature. The RRMB as shown in Fig.~\ref{fig:fig4b} is proposed for the task of image feature representation learning. To capture the features of the different scales, we used three scales, which were obtained by split the $1 * 1$ patch, $2 * 2$ patches, and $8 * 8$ patches. When splitting the $1 * 1$ patch in the IGCN $1 * 1$, which is the same with the standard convolutional layer. The scale was designed to capture global image-level features. The second scale was obtained by splitting $2 * 2$ patches in the IGCN $2 * 2$;  it was designed to ensure the stability of the features during flipped situation. This scale setting was exploited to capture the object- level features. The third scale was obtained by splitting the $8 * 8$ patches in the IGCN; the $8 * 8$ was designed such that it could construct the relationship between the relational spatial patches, such as eyes and mouth. This scale setting was exploited to capture the patch-level features. All these scales features were summed pixel-wisely to obtain the final output features.

\subsection{Loss Functions}\label{sec:subsection3-2}
\noindent
Three loss functions are mainly used in our model: pixel loss, perceptual loss, and adversarial loss. 
\begin{itemize} 
    \item \textbf{pixel loss:} The pixel loss $L_{pixel}$ used in this study is defined as follows: 
\begin{equation}
\label{eq_Lpix}
L_{pix}=\sum E_{I_{gt},I_{out}}[\left \| I_{gt}- I_{out}\right \|_2]
\end{equation}
The pixel loss constrains the generator to generate a face image. Where $I_{out}$ is the face image generated by the generator, $I_{gt}$ is the ground-truth face image. 
\end{itemize}
\begin{itemize}
    \item \textbf{perceptual loss:} We used the pre-trained 19-layer VGG to compute the perceptual
loss $L_{per}$ to obtain more facial details \cite{ledig2017photo}. The perceptual loss $L_{per}$ is defined as follows:
\begin{equation}
\label{eq_Lper}
\scalebox{.88}{$
L_{per} =\sum E_{I_{gt},I_{out}}[\left \| f_{I_{gt}}^{2,2}- f_{I_{out}}^{2,2} \right \|_2 + \left \| f_{I_{gt}}^{5,4}-f_{I_{out}}^{5,4}\right \|_2]
$}
\end{equation}
where $f_{I_{gt}}^{i,j}$ is the ground-truth’s feature map obtained by the $j$-th convolution layer before the $i$-th max-pooling layer in the VGG-19, $f_{I_{out}}^{i,j}$ is the generated face’s feature map, and $f_{I_{gt}}^{i,j}$ is the ground-truth face feature map.
\end{itemize}
\begin{itemize}
    \item \textbf{adversarial loss:} The adversarial loss $L_{adv}$ is used to constrain the generated face image to be closer to the real image; it is defined as follows:
\begin{equation}
\label{eq_Ladv}
\begin{split}
L_{adv} =&\sum E_{I_{out}\sim p(I_{out})}[log{D_I(I_{out})}]+ \\
         &\sum E_{I_{gt}\sim p(I_{gt})}[log{(1-D_I(I_{gt}))}]
\end{split}
\end{equation}
where $D_I$ is the discriminator for discriminating the ground-truth face image from the generated one.
\end{itemize}
\begin{itemize}
    \item \textbf{overall loss:} The overall loss of the proposed face image restoration framework is as follows:
\begin{equation}
L =\lambda_{1}L_{pix}+\lambda_{2}L_{adv}+\lambda_{3}L_{per}
\end{equation}
where $\lambda_1$, $\lambda_2$, $\lambda_3$ are the trade-off parameters.
\end{itemize}

\section{Experiment}

\subsection{Datasets and Settings}
\noindent
\textbf{Datasets:} We performed experimental evaluations using two public-domain datasets: \textbf{CelebA}~\cite{liu2018large}  and \textbf{Helen}~\cite{le2012interactive}. CelebA is a large-scale face attribute dataset with 10,177 subjects and 202,599 face images. We adhered to the standard protocol, and divided the dataset into a training set (162,770 images), a validation set (19,867 images), and a test set (19,962 images). Helen is composed of 2,330 face images. Based on the standard protocol of Helen, we used 2,000 images for training and 330 images for testing. In our experiments, CelebA was used to train the network, and obtain test results. Helen was used to perform cross-validation to further verify the validity of the model.

\noindent
\textbf{Implementation details:} For CelebA, we roughly aligned to 144 $*$ 144, and then randomly cropped the images to 128 $*$ 128 as inputs based on \cite{radford2015unsupervised}. For Helen, we aligned the pictures based on the 5-point coordinates detected by the multi-task cascaded convolutional networks (MTCNN) \cite{zhang2016joint}, and resized them to 128 $*$ 128 $*$ 3. For the multi-task experiments, the input ill face images were produced by resizing the high-resolution face images to 32 $* $ 32 and 16 $*$ 16 through the bicubic interpolation method, and we incorporated a random binary mask whose size is one-fourth of the input size. For the face completion experiment, a mask concealing a quarter of the full image was added to the input image; for the face super-resolution, we resized the high-resolution face image to 16 $*$ 16 through the bicubic interpolation method to conduct the experiment with the eight-times downsampled image. We used an Adam algorithm with an initial learning rate of $10^{-4}$ to optimize the face image restoration network. The settings of the trade-off parameters were $\lambda_1$=1, $\lambda_2$=0.01, and $\lambda_3$=0.0005. The batch size was 24, and the kernel size was 3.

\noindent
\textbf{Evaluation metrics:} We evaluated the model mainly from the qualitative and quantitative aspects.

\begin{itemize}
    \item \textbf{Qualitative evaluation metrics:} We evaluated the images based on multi-task face super-resolution and face completion, and conducted an ablation study to observe the quality of the repair. 
\end{itemize}

\begin{itemize}
    \item \textbf{Quantitative evaluation metrics:} We quantitatively evaluated the repaired images using two main metrics, the \textbf{(PSNR)} and the structural similarity index \textbf{(SSIM)}~\cite{wang2004image}.
\end{itemize}

\subsection{Quality results}
\begin{figure*}[htb] 
    \centering
    \subfloat[Input: 8-times downsampled image with $1/4$ area of occlusion]{%
        \includegraphics[width=0.45\textwidth]{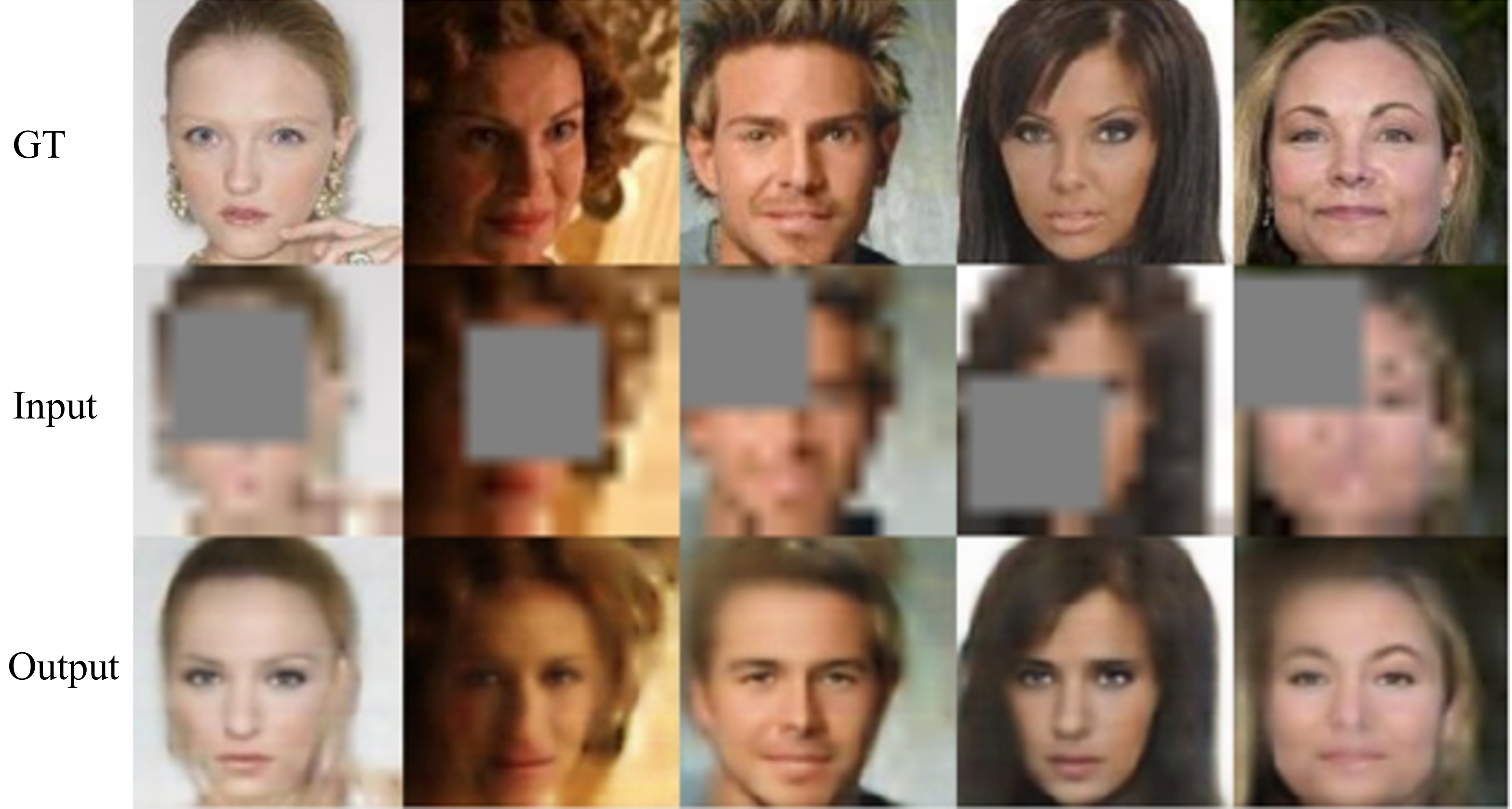}%
        \label{fig:fig5a}%
        }%
    \hfill%
    \subfloat[Input: 4-times downsampled image with $1/4$ area of occlusion]{%
        \includegraphics[width=0.45\textwidth]{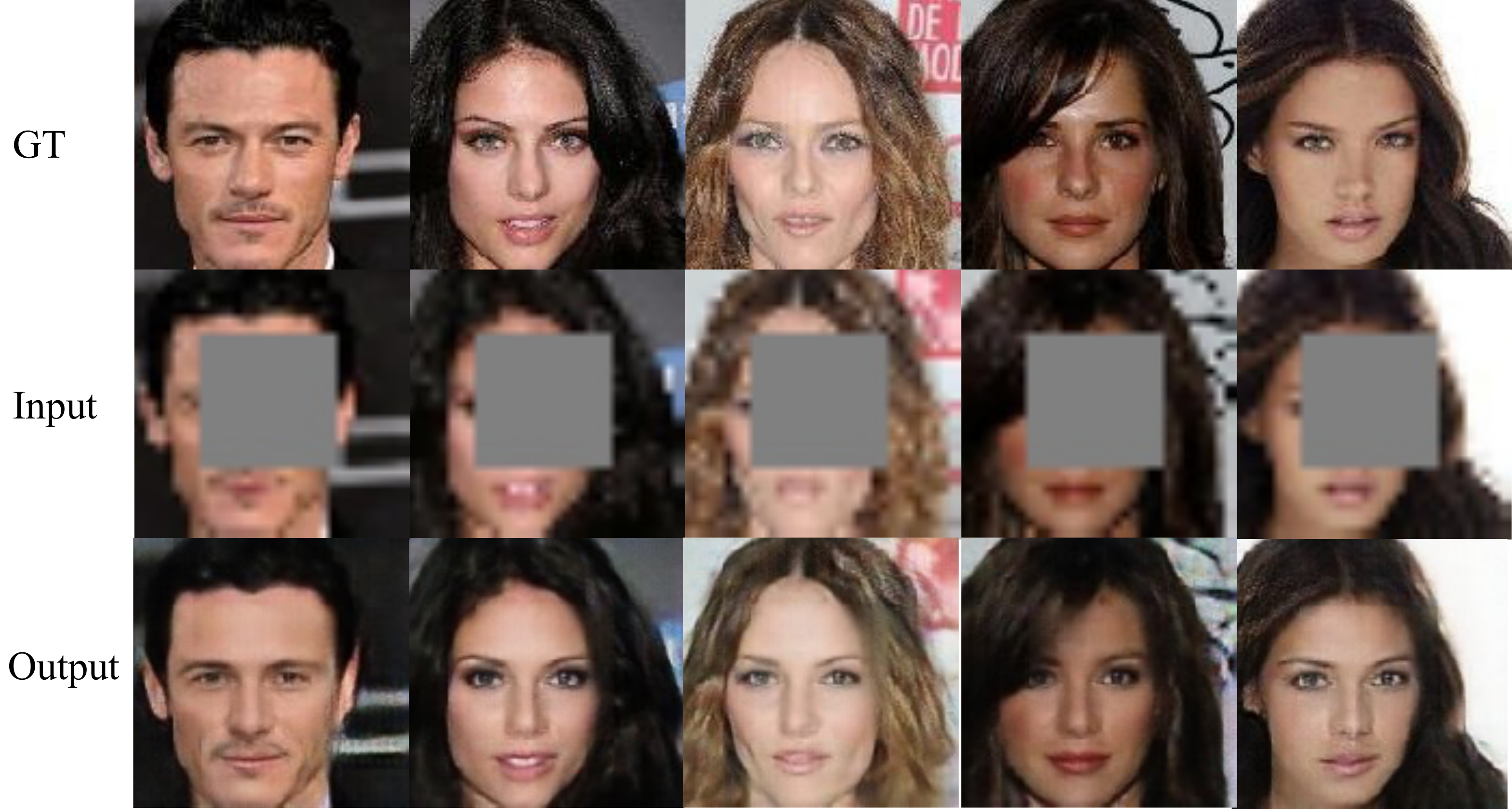}%
        \label{fig:fig5b}%
        }%
    \caption{Qualitative results on multi-task experiment under different input conditions. The first row is the real picture, the second row is the input, and the last is the output.}
    \label{fig:fig5}
\end{figure*}

\noindent
Qualitative results can provide an intuitive observation of the recovered face images through different methods. For joint face completion and face super-resolution, we performed two experiments (SRFC x4 and SRFC x8) with the CelebA dataset as the input. We used a four-times downsampled image(from 128 $*$ 128 to 32 $*$ 32) with 1/4 area of occlusion. The input was an eight-times downsampled image(from 128 $*$ 128 to 16 $*$ 16), with 1/4 area of occlusion. The qualitative results are shown in Fig.~\ref{fig:fig5a} and Fig.~\ref{fig:fig5b}.

\begin{figure}[htb]
\centering
\includegraphics[scale=0.35]{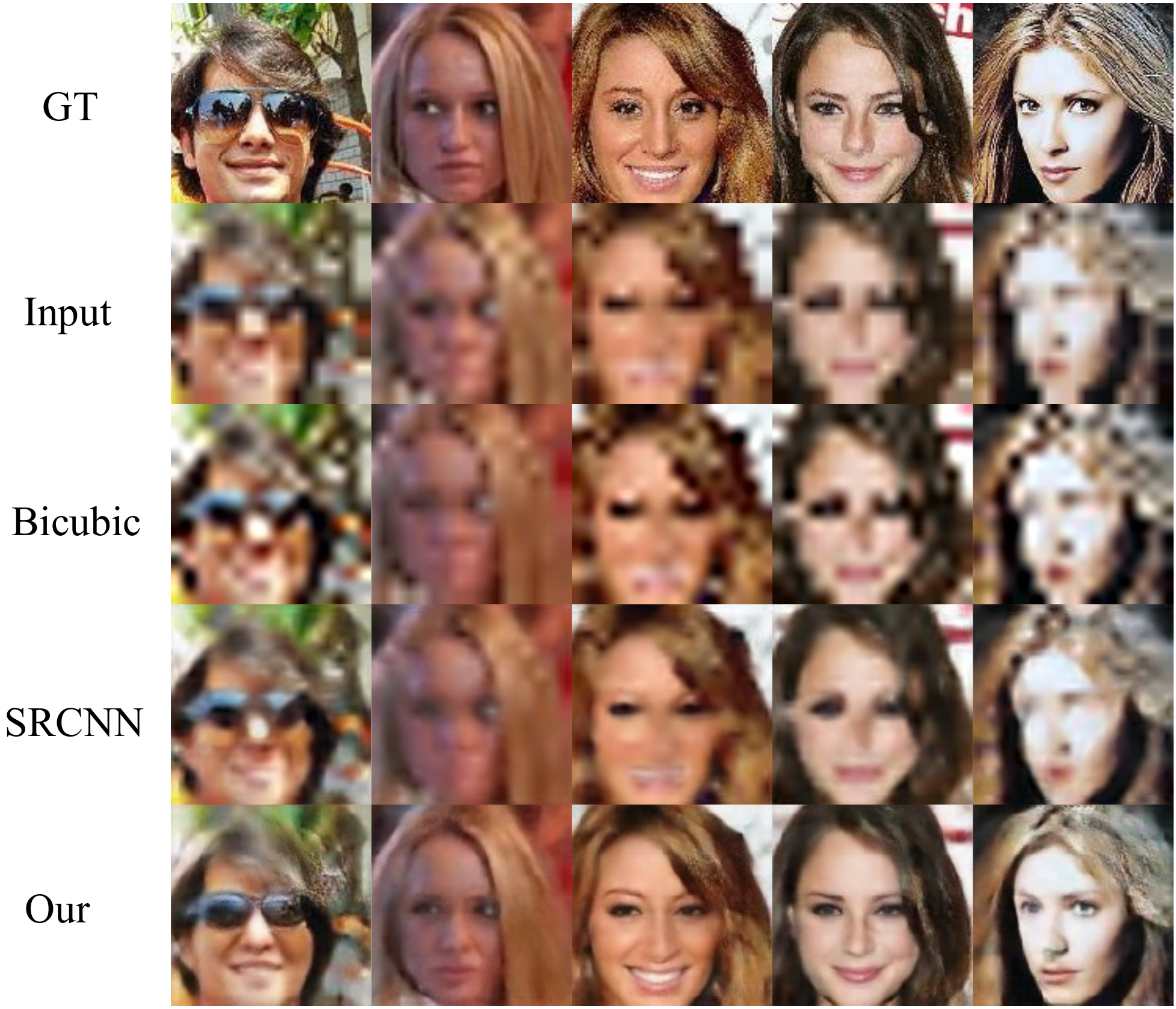}
\caption{Visualization of face super-resolution results on CelebA. Our super-resolution results vastly outperformed other methods in terms of visual quality.}
\label{fig:fig6}
\end{figure}

\begin{figure*}[htb]
\centering
\includegraphics[scale=0.4]{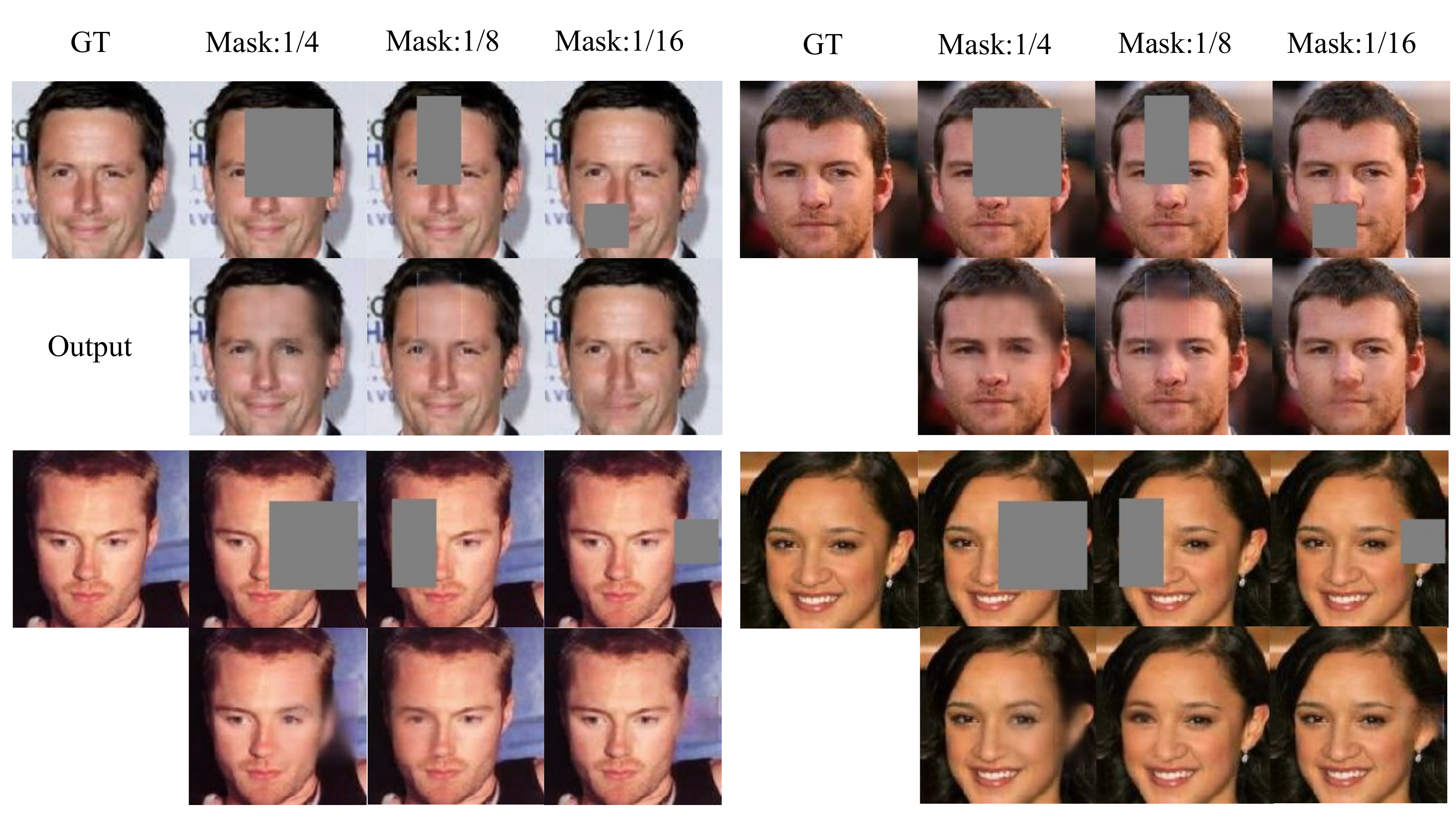}
\caption{Visualization of face completion results on CelebA. From left to right are the ground truth, and the input and output results obtained for the images with the 1/4 mask, 1/8 mask, and the 1/16 mask, respectively.}
\label{fig:fig7}
\end{figure*}

To verify the effectiveness of the model on single tasks, we compared it with two baseline methods, Bicubic and SRCNN, on face image super-resolution \cite{ren2016single}. The comparison results are shown in Fig.~\ref{fig:fig6}. In addition, we also verified the robustness of the model to variations in the size of the occlusion toward face completion; we set the size of the mask to 1/4, 1/8, and 1/16 of the original image, and observed the results obtained for the different mask sizes, as shown in Fig.~\ref{fig:fig7}.

It can be observed that the smaller the occluded area, the better the restored image’s quality. This also shows that the model is robust for face completion.

\subsection{Quantity results}
\noindent
In addition to visual quality, we also quantified the effectiveness of the proposed approach at face completion and super-resolution based on two measurements. One is the PSNR, which is widely used during image compression to measure the fidelity of the reconstructed signal w.r.t. ground-truth. The other is the SSIM \cite{wang2004image}, a perceptual measure that considers image degradation based on several perceptual information, such as luminance and contrast, in addition to the perceived change in structural information.

\begin{table}[!htbp]
\begin{center}
\caption{Quantitative results for joint face completion and super-resolution. ``SRFC x4'' represents the four-times downsampled input image with a 1/4 area of occlusion; "SRFC x8" represents the eight-times downsampled input image with a 1/4 area of occlusion. The red type indicates the best performance.}\label{tab:tab1}
\begin{tabular}{c||c|c|c}
\hline
\textbf{SRFC} & \textbf{SRFC x8} & \textbf{FCSR-GAN x4} & \textbf{SRFC x4}    \\ \hline\hline
\textbf{PSNR} & 21.19            & 22.23                & {\textcolor{red}{23.49}} \\ \hline
\textbf{SSIM} & 0.634            & 0.657                & {\textcolor{red}{0.714}} \\ \hline
\end{tabular}
\end{center}
\end{table}

The quantitative results for the joint face completion and super-resolution is shown in Table~\ref{tab:tab1}, where ``SRFC x4'' represents the four-times downsampled input image with $1/4$ area of occlusion; it can be observed that the proposed model outperforms the FCSR-GAN based on both the PSNR and SSIM indicators. The ``SRFC x8'' represents the eight-times downsampled image with $1/4$ area of occlusion. We conducted a cross-dataset validation to evaluate the generalizability of our MFG-GAN, i.e., the model was trained on CelebA, but tested on Helen. Then, the model pre-trained using CelebA was fine-tuned and tested with Helen. All through, the inputs were $32 * 32$ face images with $1/4$ area of occlusion. The result is shown in Table~\ref{tab:tab2}. 

\begin{table}[!htbp]
\begin{center}
\caption{The quantitative results for joint face completion and super-resolution on Helen. ``train'' indicates that the model was trained on CelebA, and testing was conducted on Helen. ``fine-tune'' indicates that the model was trained on CelebA, and fine-tuned on Helen.}\label{tab:tab2}
\begin{tabular}{c||c|c}
\hline
\textbf{SRFC}      & \textbf{train} & \textbf{fin-tune} \\ \hline\hline
\textbf{PSNR} & 20.772            & 22.442              \\ \hline
\textbf{SSIM} & 0.639             & 0.692               \\ \hline
\end{tabular}
\end{center}
\end{table}

Our MFG-GAN trained on CelebA achieved 20.772 dB PSNR and 0.639 SSIM on Helen; and fine-tuning it on Helen achieved 22.442 dB PSNR and 0.692 SSIM. Compared with the intra-database testing results on CelebA (23.49 dB PSNR and 0.714 SSIM), these results are quite encouraging, considering the difference of the data distributions between CelebA and Helen.

For face completion, we compared the proposed model with the two baseline models, CE~\cite{pathak2016context} and GFC~\cite{li2017generative}. We ensured that the inputs in all the instances were $128 * 128$ face images with 1/4 area of occlusion, as shown in Table~\ref{tab:tab3}. For the face super-resolution, we compared the proposed model with two baseline models, Bicubic and SRCNN~\cite{ren2016single}, and ensured that the inputs in all the instances were $16 * 16$ face images, as shown in Table~\ref{tab:tab4}. 

\begin{table}[!htbp]
\begin{center}
\caption{Quantitative results of face completion on CelebA testing sets. Red type indicates the best performance.}\label{tab:tab3}
\begin{tabular}{c||c|c|c}
\hline
\textbf{FC}  & \textbf{CE~\cite{pathak2016context}} & \textbf{GFC~\cite{li2017generative}} & \textbf{our}    \\ \hline\hline
\textbf{PSNR} & 24.499            & 24.281                & {\textcolor{red}{25.413}} \\ \hline
\textbf{SSIM} & 0.732            & 0.837                & {\textcolor{red}{0.861}} \\ \hline
\end{tabular}
\end{center}
\end{table}

\begin{table}[!htbp]
\begin{center}
\caption{Quantitative results of face super-resolution on CelebA testing sets. Red type indicates the best performance.}\label{tab:tab4}
\begin{tabular}{c||c|c|c}
\hline
\textbf{SR}  & \textbf{Bicubic} & \textbf{SRCNN~\cite{ren2016single}} & \textbf{our}    \\ \hline\hline
\textbf{PSNR} & 21.049  & 21.938  & {\textcolor{red}{23.307}} \\ \hline
\textbf{SSIM} & 0.601   & 0.632   & {\textcolor{red}{0.701}} \\ \hline
\end{tabular}
\end{center}
\end{table}

Furthermore, we verified the robustness of the model to variations in the occlusion size, as shown in Table~\ref{tab:tab5}, where "mask:1/4" corresponds to $128 * 128$ input face images with $1/4$ area of occlusion; "mask:1/8" corresponds to $128 * 128$ input face images with $1/8$ area of occlusion, and "mask:1/16" corresponds to $128 * 128$ face images with $1/16$ area of occlusion.

\begin{table}[!htbp]
\begin{center}
\caption{Quantitative results of face completion on CelebA testing sets with different occlusions' sizes.}\label{tab:tab5}
\begin{tabular}{c||c|c|c}
\hline
\textbf{FC}  & \textbf{mask:1/4} & \textbf{mask:1/8} & \textbf{mask:1/16}    \\ \hline\hline
\textbf{PSNR} & 25.413& 28.891& 33.417 \\ \hline
\textbf{SSIM} & 0.861& 0.922& 0.962 \\ \hline
\end{tabular}
\end{center}
\end{table}

It can be observed that as the size of the occlusion became smaller, the repair effect improved, which shows that the model demonstrates some robustness in face completion.

\subsection{Ablation Study}
\noindent
The proposed MFG-GAN consists of the IGCN and FPN. We designed three models to verify the effectiveness of both part: \textbf{M1}, \textbf{M2}, \textbf{M3}. The FPN is absent in \textbf{M1}; general convolution is used in place of graph convolution in \textbf{M2}. In \textbf{M3}, the IGCN and FPN are integrated. 

\begin{figure}[htbp]
\centering
\includegraphics[scale=0.4]{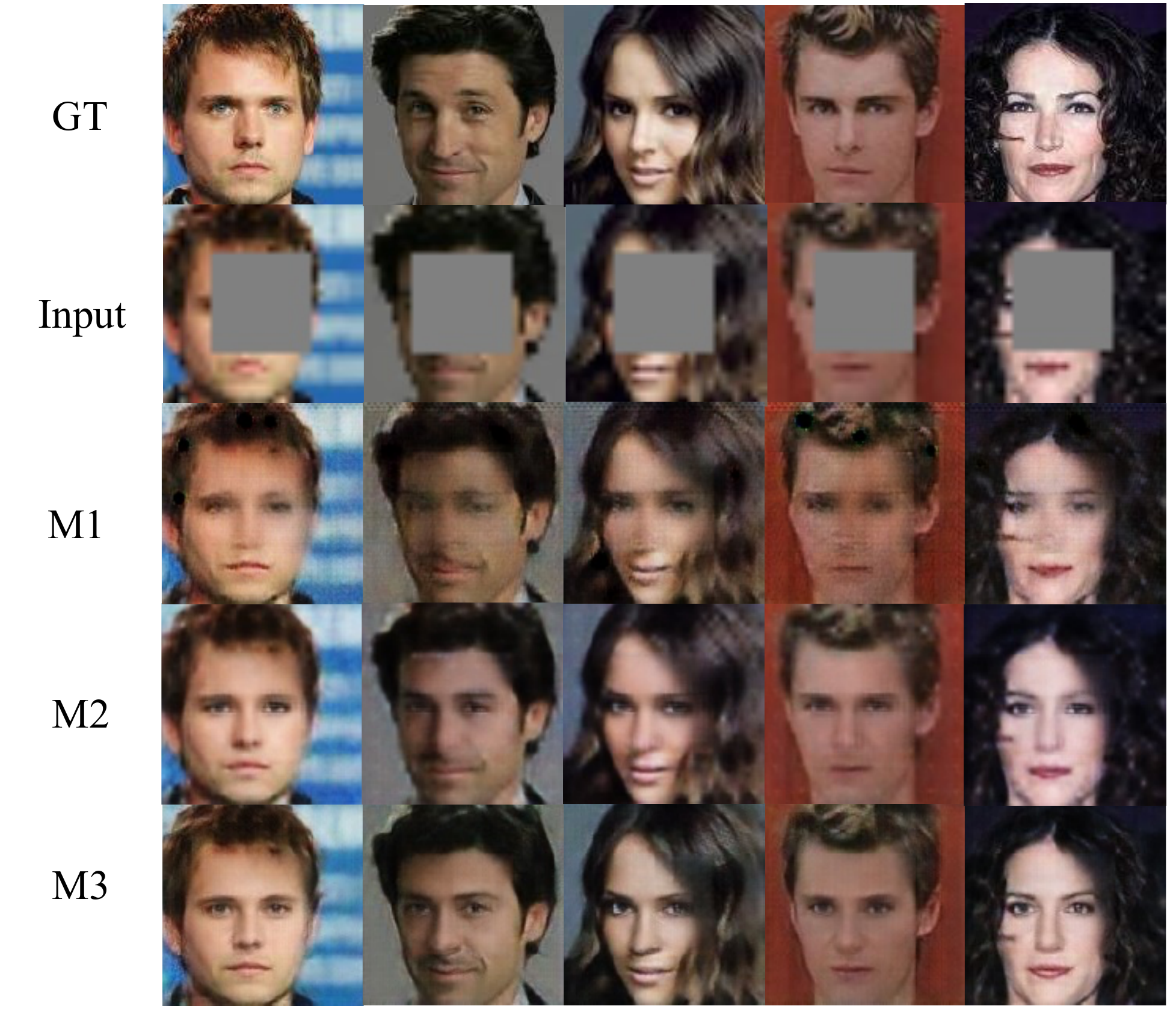}
\caption{Visualization of the ablation study. The first two columns are the real image and input respectively. The third column shows the output results of the model without the FPN. The fourth column shows the output results when graph convolution is replaced with conventional convolution; the last column is the output of the experimental model. It can be observed that the model combining both the IGCN and FPN achieves the best qualitative results.}
\label{fig:fig8}
\end{figure}

\begin{table}[!htbp]
\begin{center}
\caption{Quantitative results of ablation study, where ``M1'' is the model without the FPN, ``M2'' uses general convolution instead of graph convolution, and ``M3'' is the combination of the IGCN and FPN. Red type indicates the best performance.}\label{tab:tab6}
\begin{tabular}{c||c|c|c|c}
\hline
\textbf{SRFC} & \textbf{FPN} & \textbf{IGCN} & \textbf{PSNR} & \textbf{SSIM}    \\ \hline\hline
\textbf{M1}& &  \checkmark  & 21.544  & 0.624 \\ \hline
\textbf{M2}&  \checkmark&   & 22.569    & 0.687\\ \hline
\textbf{M3}& \checkmark& \checkmark & \textcolor{red}{23.499}&\textcolor{red}{0.714}\\ \hline
\end{tabular}
\end{center}
\end{table}

All the experiments use the same input: four-times downsampled image (i.e., from $128 * 128$ to $32 * 32$) with 1/4 area of occlusion. The results are shown in Table~\ref{tab:tab6}. The qualitative results obtained using these three models with CelebA are shown in Fig.~\ref{fig:fig8}. In addition, to demonstrate the impact of the IGCN and FPN on the restored results more intuitively, we show the results from the details. 

\begin{figure}[htb]
\centering
\includegraphics[scale=0.4]{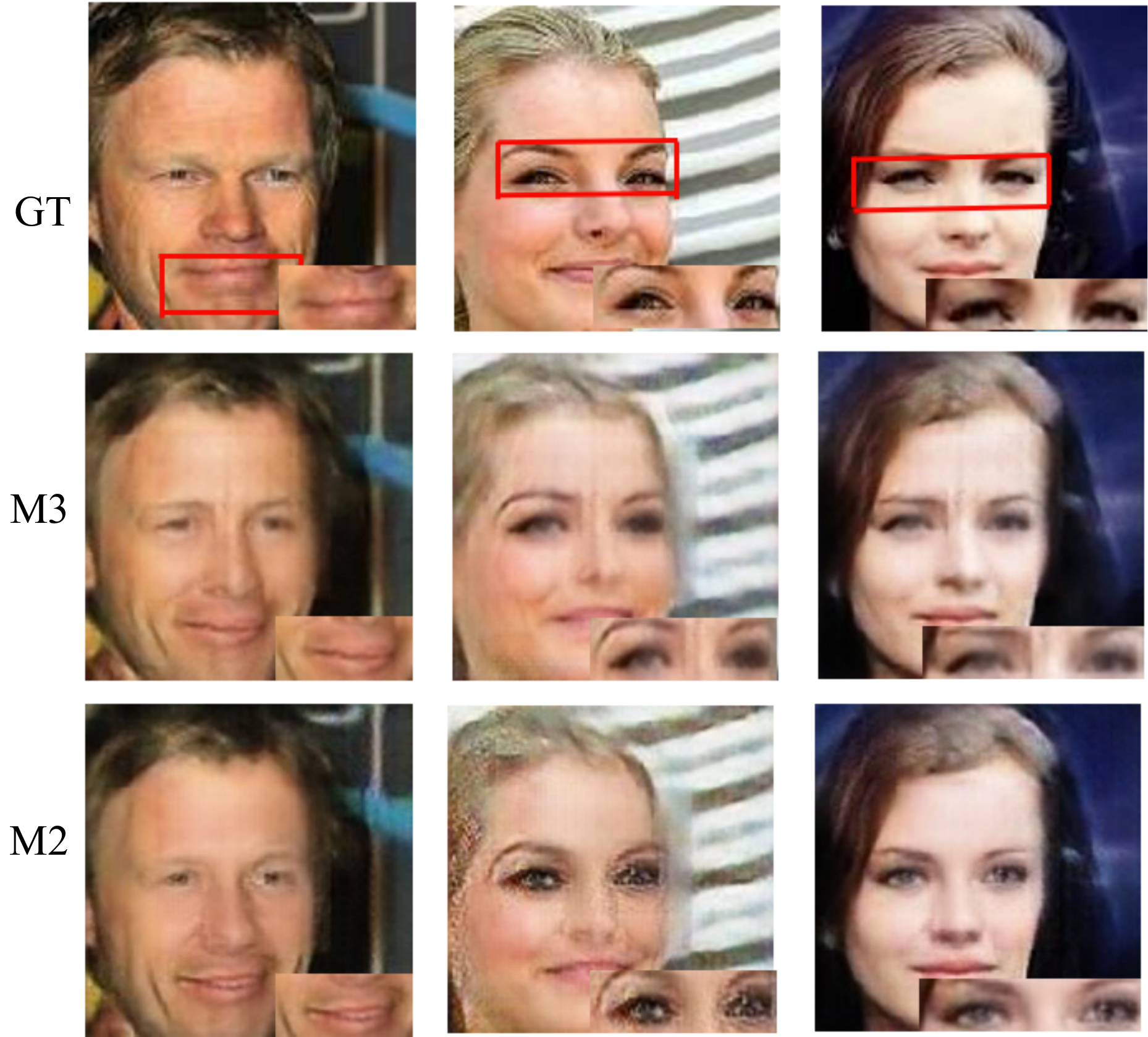}
\caption{Comparison results of face image restoration method with and without IGCN. By making it possible to learn the correlation among the various regions of the face, the IGCN made the restoration of facial features more accurate.}
\label{fig:fig9}
\end{figure}

\begin{figure}[htb]
\centering
\includegraphics[scale=0.35]{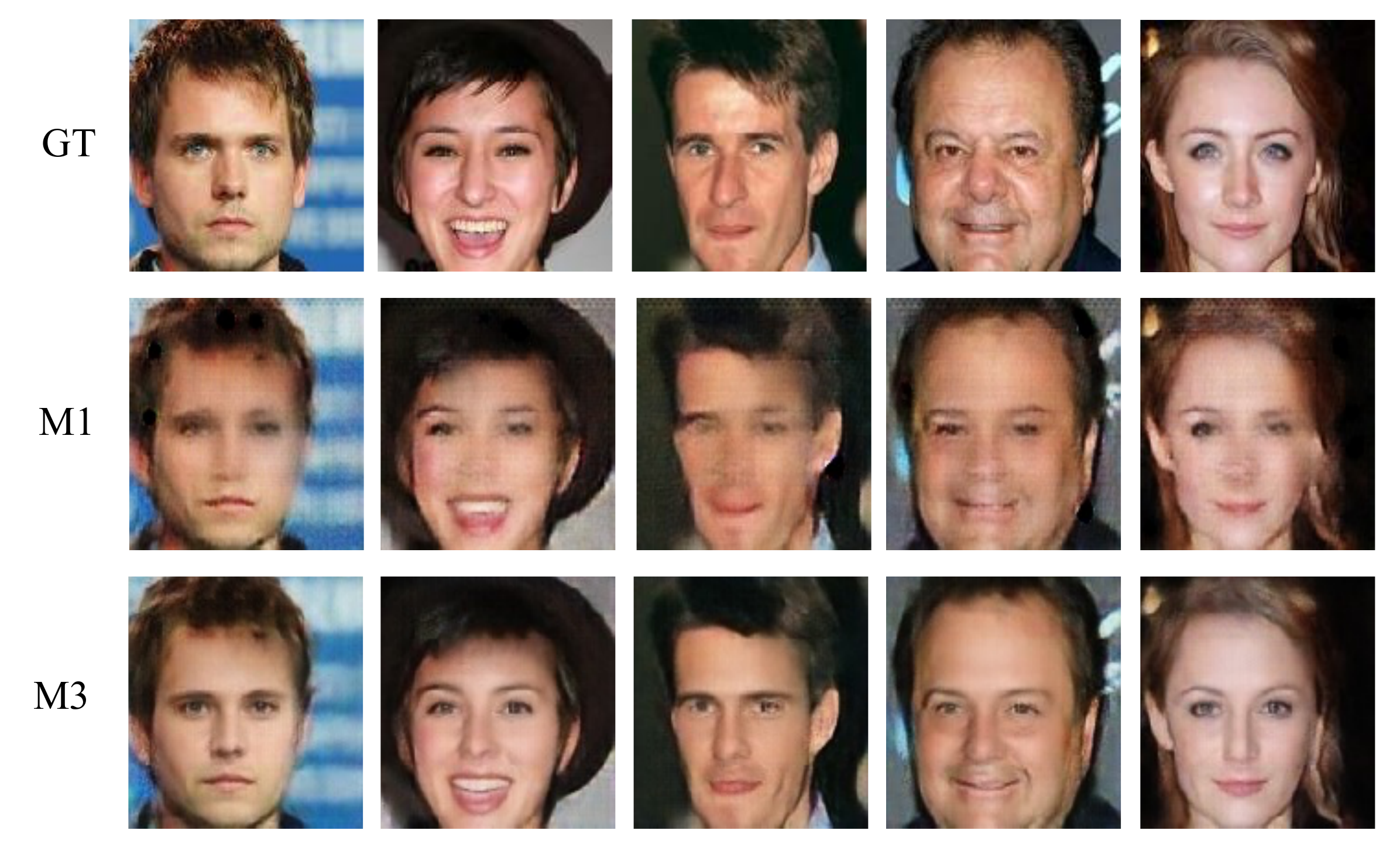}
\caption{Comparison of image restoration with and without FPN. Incorporating the FPN improved the repair results of some small facial features such as eyes and nose.}
\label{fig:fig10}
\end{figure}

Fig.~\ref{fig:fig9} and Fig.~\ref{fig:fig10} show the impact of the IGCN and the FPN on the restoration results, respectively. Comparing the results of the two images, it is apparent that because the IGCN takes into account the relationship between features, the detailed information learned in some samples was more accurate; furthermore, the extraction of the multi-scale features by the FPN also remarkably improves the learning of some small-sized features such as eyes and nose.

\begin{figure}[htb]
\centering
\includegraphics[scale=0.35]{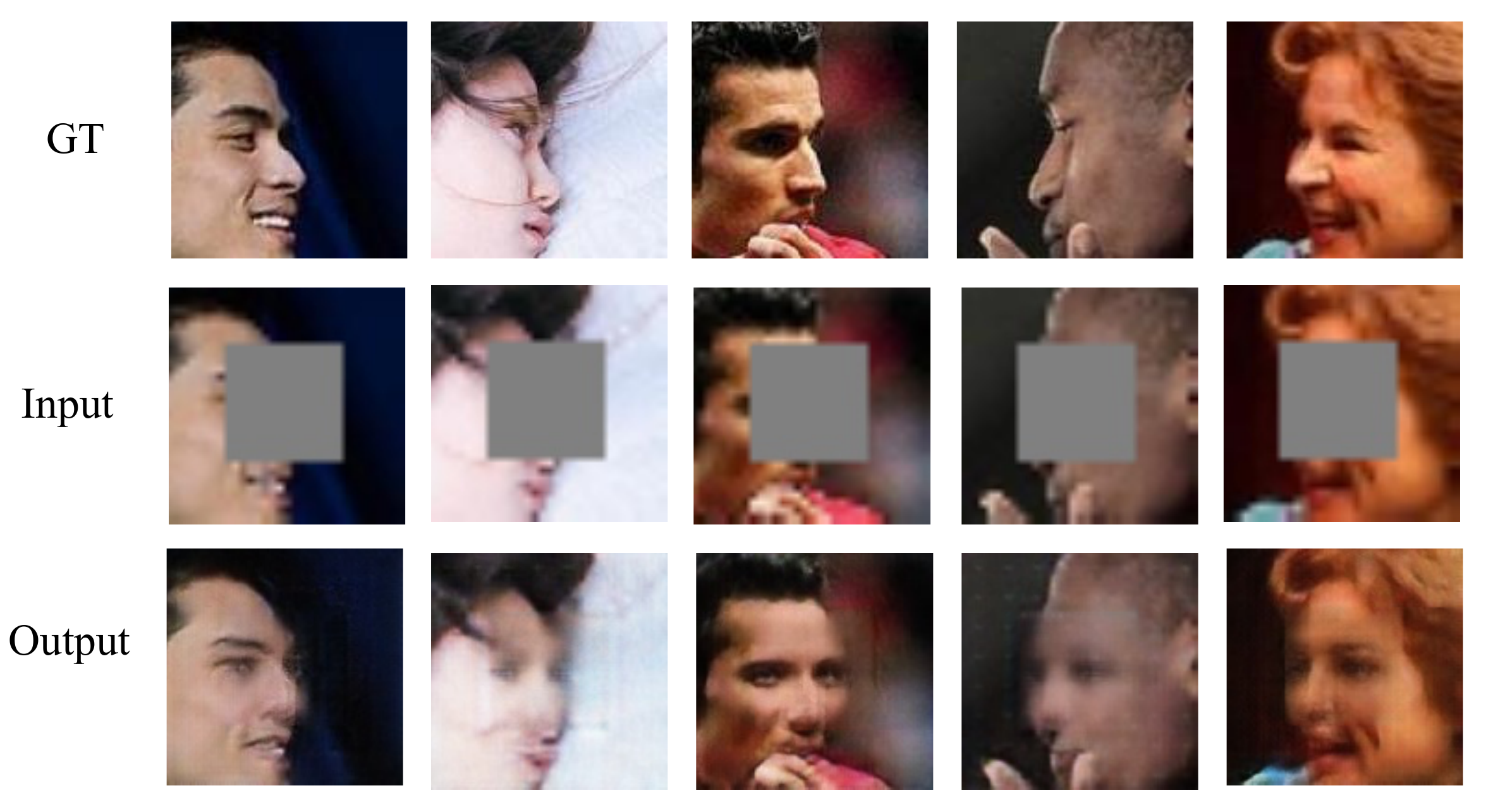}
\caption{Unrealistic restoration results for special faces. Due to the lack of facial information, the relationship between the repaired area and the complete area cannot be learned, resulting in the inability to restore the original effect.}
\label{fig:fig11}
\end{figure}

We also list some unrealistic repair results. As shown in Fig.~\ref{fig:fig11}, it can be found that these pictures are mostly in profile. This may be attributable to the impossibility of fully learning the correlation among the facial features due to the special nature of the data; thus, the repair is ineffective.

\section{Conclusion}
\noindent
In this paper, based on the integration of an IGCN and FPN, we proposed a joint face completion and face super-resolution method (MFG-GAN) that can leverage multi-task learning to recover non-occluded high-resolution facial patches from LR face images with occlusions. The experimental results on the public datasets, CelebA and Helen, reveal that the proposed model outperforms the baseline method when simultaneously tackling the tasks of face completion and face image super-resolution. Furthermore, the proposed method proved to be effective following both cross-dataset and internal dataset testing. In addition, we verified the model’s effectiveness on the tasks of face completion and face super-resolution singly, and achieved outstanding results. The proposed framework  also has prospects for other face restoration tasks and other multi-task problems such as face recognition and facial attribute analysis.

The proposed method can mainly be used to repair occluded and LR face images. The evaluation indicators were the most commonly used indicators related to image quality evaluation (PSNR and SSIM). In future work, we shall further evaluate the repair results based on the identity information of the face, and attempt to retain it as much as possible. Furthermore, we will also exploit some prior knowledge of faces to optimize the repair results.

% if have a single appendix:
%\appendix[Proof of the Zonklar Equations]
% or
%\appendix  % for no appendix heading
% do not use \section anymore after \appendix, only \section*
% is possibly needed

% use appendices with more than one appendix
% then use \section to start each appendix
% you must declare a \section before using any
% \subsection or using \label (\appendices by itself
% starts a section numbered zero.)
%

%\appendices
%\section{Proof of the First Zonklar Equation}
%Appendix one text goes here.
%
%% you can choose not to have a title for an appendix
%% if you want by leaving the argument blank
%\section{}
%Appendix two text goes here.

% use section* for acknowledgment
\ifCLASSOPTIONcompsoc
  % The Computer Society usually uses the plural form
  \section*{Acknowledgments}
\else
  % regular IEEE prefers the singular form
  \section*{Acknowledgment}
\fi

This work is supported by National Natural Science Foundation of China ( No.41806116 and No.61503277 ). We gratefully acknowledge the support of NVIDIA Corporation with the donation of the Titan V GPU used for this research.
% Can use something like this to put references on a page
% by themselves when using endfloat and the captionsoff option.
\ifCLASSOPTIONcaptionsoff
  \newpage
\fi

% trigger a \newpage just before the given reference
% number - used to balance the columns on the last page
% adjust value as needed - may need to be readjusted if
% the document is modified later
%\IEEEtriggeratref{8}
% The "triggered" command can be changed if desired:
%\IEEEtriggercmd{\enlargethispage{-5in}}

% references section

% can use a bibliography generated by BibTeX as a .bbl file
% BibTeX documentation can be easily obtained at:
% http://mirror.ctan.org/biblio/bibtex/contrib/doc/
% The IEEEtran BibTeX style support page is at:
% http://www.michaelshell.org/tex/ieeetran/bibtex/
\bibliographystyle{IEEEtran}
% argument is your BibTeX string definitions and bibliography database(s)
\bibliography{refs}

% or if you just want to reserve a space for a photo:

%\begin{IEEEbiography}{Michael Shell}
%Biography text here.
%\end{IEEEbiography}
%
%% if you will not have a photo at all:
%\begin{IEEEbiographynophoto}{John Doe}
%Biography text here.
%\end{IEEEbiographynophoto}
%
%% insert where needed to balance the two columns on the last page with
%% biographies
%%\newpage
%
%\begin{IEEEbiographynophoto}{Jane Doe}
%Biography text here.
%\end{IEEEbiographynophoto}

% You can push biographies down or up by placing
% a \vfill before or after them. The appropriate
% use of \vfill depends on what kind of text is
% on the last page and whether or not the columns
% are being equalized.

%\vfill

% Can be used to pull up biographies so that the bottom of the last one
% is flush with the other column.
%\enlargethispage{-5in}

% that's all folks
\end{document}